\documentclass[10pt,twocolumn,letterpaper]{article}

\usepackage{cvpr}              
\usepackage[accsupp]{axessibility}
\usepackage{graphicx}
\usepackage{amsmath}
\usepackage{amssymb}
\usepackage{booktabs}
\usepackage{float}

\usepackage[pagebackref,breaklinks,colorlinks]{hyperref}
\usepackage[normalem]{ulem}
\useunder{\uline}{\ul}{}

\usepackage[capitalize]{cleveref}
\crefname{section}{Sec.}{Secs.}
\Crefname{section}{Section}{Sections}
\Crefname{table}{Table}{Tables}
\crefname{table}{Tab.}{Tabs.}

\def\cvprPaperID{10496} 
\def\confName{CVPR}
\def\confYear{2023}

\begin{document}

\title{Temporal Interpolation Is All You Need for Dynamic Neural Radiance Fields}

\author{Sungheon Park,\, Minjung Son,\, Seokhwan Jang,\, Young Chun Ahn,\, Ji-Yeon Kim,\, Nahyup Kang\\
Samsung Advanced Institute of Technology (SAIT)\\
{\tt\small \{sh2019.park, minjungs.son, swan.jang, ychun.ahn, jiyeon31.kim, nahyup.kang\}@samsung.com}
}

\makeatletter
\g@addto@macro\@maketitle{
  \begin{figure}[H]
  \centering
  \includegraphics[width=0.95\textwidth]{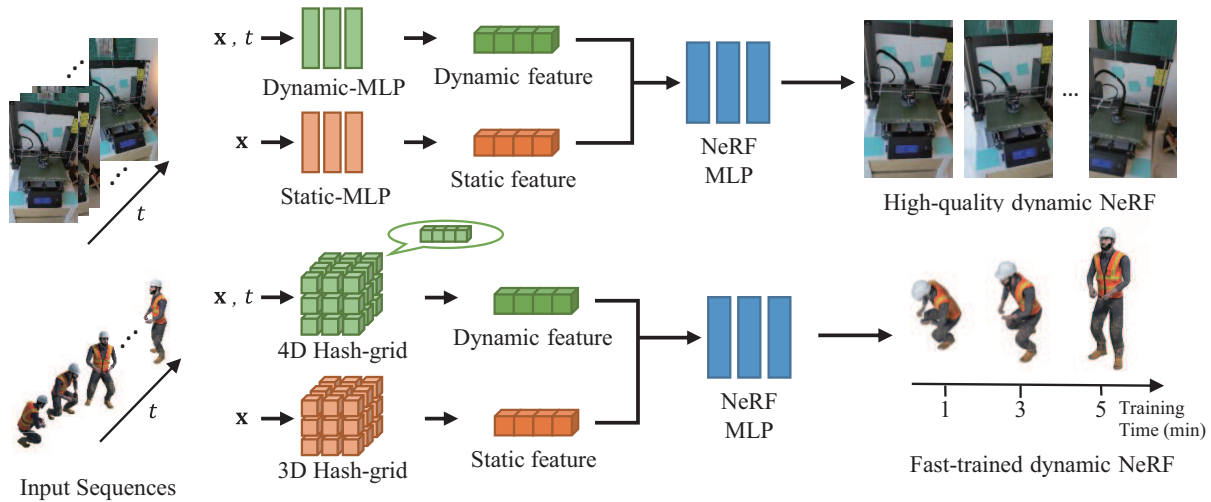}
  \setlength{\linewidth}{\textwidth}
  \setlength{\hsize}{\textwidth}
  \caption{We propose simple yet effective feature interpolation methods for training neural radiance fields of dynamic scenes based on temporal interpolation. We provide two different feature vector representations, neural representation (top) and grid representation (bottom), both of which are the concatenation of static feature vectors and temporally-interpolated dynamic feature vectors. The neural representation exhibits high-quality rendering performance with small-sized models while the grid representation shows competitive rendering results with astonishingly fast training speed.}
  \label{fig:intro}
  \end{figure}
}
\makeatother

\maketitle

\begin{abstract}
Temporal interpolation often plays a crucial role to learn meaningful representations in dynamic scenes. In this paper, we propose a novel method to train spatiotemporal neural radiance fields of dynamic scenes based on temporal interpolation of feature vectors. Two feature interpolation methods are suggested depending on underlying representations, neural networks or grids. In the neural representation, we extract features from space-time inputs via multiple neural network modules and interpolate them based on time frames. The proposed multi-level feature interpolation network effectively captures features of both short-term and long-term time ranges. In the grid representation, space-time features are learned via four-dimensional hash grids, which remarkably reduces training time. The grid representation shows more than 100$\times$ faster training speed than the previous neural-net-based methods while maintaining the rendering quality. Concatenating static and dynamic features and adding a simple smoothness term further improve the performance of our proposed models. Despite the simplicity of the model architectures, our method achieved state-of-the-art performance both in rendering quality for the neural representation and in training speed for the grid representation.
\end{abstract}


\section{Introduction}
\label{sec:intro}
3D reconstruction and photo-realistic rendering have been long-lasting problems in the fields of computer vision and graphics. Along with the advancements of deep learning, differentiable rendering~\cite{kato2018neural, liu2019soft} or neural rendering, has emerged to bridge the gap between the two problems. Recently proposed Neural Radiance Field (NeRF)~\cite{mildenhall2021nerf} has finally unleashed the era of neural rendering. Using NeRF, it is able to reconstruct accurate 3D structures from multiple 2D images and to synthesize photo-realistic images from unseen viewpoints. Tiny neural networks are sufficient to save and retrieve complex 3D scenes, which can be trained in a self-supervised manner given 2D images and camera parameters.

Meanwhile, as our world typically involves dynamic changes, it is crucial to reconstruct the captured scene through 4D spatiotemporal space. Since it is often not possible to capture scenes at different viewpoints simultaneously, reconstructing dynamic scenes from images is inherently an under-constrained problem.
While NeRF was originally designed to deal with only static scenes, there have been a few approaches in the literature that extend NeRF to dynamic scenes~\cite{park2021hypernerf,pumarola2021d,tretschk2021non,li2021neural}, which are so called as dynamic NeRFs. Inspired by the non-rigid structure-from-motion algorithms~\cite{bregler2000recovering, akhter2010trajectory} that reconstruct 3D structures of deformable objects, most previous works solved the under-constrained setting by estimating scene deformations~\cite{park2021nerfies,park2021hypernerf,tretschk2021non} or 3D scene flows~\cite{li2022neural,gao2021dynamic,du2021neural} for each frame.

However, since the parameters of deformation estimation modules are jointly optimized with NeRF network simultaneously, it is questionable that the modules can accurately estimate deformations or scene flows in accordance with its design principles. In many dynamic scenes, it is challenging to resolve the ambiguities whether a point was newly appeared, moved, or changed its color. It is expected that those ambiguities and deformation estimation can be separately solved within a single network, but in practice, it is hard to let the network implicitly learn the separation, especially for general dynamic scenes without any prior knowledge of deformation.

On the other hand, grid representations in NeRF training~\cite{sun2022direct,fridovich2022plenoxels,mueller2022instant} have grabbed a lot of attentions mainly due to its fast training speed. Simple trilinear interpolation is enough to fill in the 3D space between grid points. While the representation can be directly adopted to dynamic NeRFs together with the warping estimation module~\cite{fang2022fast}, it still requires additional neural networks that affects training and rendering speed. 

Motivated by the aforementioned analyses, we present a simple yet effective architecture for training dynamic NeRFs. The key idea in this paper is to apply feature interpolation to the temporal domain instead of using warping functions or 3D scene flows. While the feature interpolation in 2D or 3D space has been thoroughly studied, to the best of our knowledge, feature interpolation method in temporal domain for dynamic NeRF has not been proposed yet. We propose two multi-level feature interpolation methods depending on feature representation which is either neural nets or hash grids~\cite{mueller2022instant}. Overview of the two representations, namely the neural representation and the grid representation, are illustrated in \cref{fig:intro}. 
In addition, noting that 3D shapes deform smoothly over time in dynamic scenes, we additionally introduced a simple smoothness term that encourages feature similarity between the adjacent frames. We let the neural networks or the feature grids to learn meaningful representations implicitly without imposing any constraints or restrictions but the smoothness regularizer, which grants the flexibility to deal with various types of deformations. Extensive experiments on both synthetic and real-world datasets validate the effectiveness of the proposed method. We summarized the main contributions of the proposed method as follows: 


\begin{itemize}
\item We propose a simple yet effective feature extraction network that interpolates two feature vectors along the temporal axis. The proposed interpolation scheme outperforms existing methods without having a deformation or flow estimation module.
\item We integrate temporal interpolation into hash-grid representation~\cite{mueller2022instant}, which remarkably accelerates training speed more than $100\times$ faster compared to the neural network models.
\item We propose a smoothness regularizer which effectively improves the overall performance of dynamic NeRFs.
\end{itemize}

\section{Related Work}
{\bf \noindent Dynamic NeRF.} There have been various attempts to extend NeRF to dynamic scenes. 
Existing methods can be categorized into three types: warping-based, flow-based, and direct inference from space-time inputs.

Warping-based methods learn how 3D structure of the scene is deformed. 3D radiance field for each frame is warped to single or multiple canonical frames using the estimated deformation. Deformation is parameterized as 3D translation~\cite{pumarola2021d}, rotation and translation via angle-axis representation~\cite{park2021nerfies,park2021hypernerf}, or weighted translation~\cite{tretschk2021non}.

On the other hand, flow-based methods estimate the correspondence of 3d points in consecutive frames. Neural scene flow fields~\cite{li2021neural} estimates 3D scene flow between the radiance fields of two time stamps. Xian \etal~\cite{xian2021space} suggested irradiance fields which spans over 4D space-time fields with a few constraints derived from prior knowledge.

Although warping-based and flow-based approaches showed successful results, they have a few shortcomings. For instance, warping-based methods cannot deal with topological variations. Since those methods warp every input frame to a single canonical scene, it is hard to correctly represent newly appeared or disappeared 3D radiance fields in the canonical scene. HyperNeRF~\cite{park2021hypernerf} learns hyperspace which represents multidimensional canonical shape bases. Neural scene flow fields~\cite{li2021neural} solve the problem by introducing occlusion masks which are used to estimate the regions where scene flows are not applicable. However, the masks and the flow fields should be learned simultaneously during the training, which makes the training procedure complicated and imposes dependence on extra information such as monocular depth or optical flow estimation.

Without estimating shape deformations or scene flows, DyNeRF~\cite{li2022neural} used a simple neural network to train NeRFs of dynamic scenes, although the experiments are only conducted on synchronized multi-view videos. It utilizes 3D position and time frame as inputs, and predicts color and density through a series of fully connected neural networks. Our neural representation applies temporal interpolation to intermediate features which enhances representation power for dynamic features while keeping the simple structure.

Category-specific NeRFs are one of the popular research directions for training dynamic objects such as humans~\cite{Weng_2022_CVPR, gafni2021dynamic} or animals~\cite{yang2022banmo}. They are based on parametric models or use additional inputs such as pose information. Our method focuses on training NeRFs of general dynamic scenes without any prior knowledge about scenes or objects. Nevertheless, temporal interpolation proposed in this paper can be easily applied to the dynamic NeRFs with templates.

{\bf \noindent Grid representations in NeRF.} One of the major drawbacks of the original NeRF~\cite{mildenhall2021nerf} is slow training speed. A few concurrent works are proposed to speed up NeRF training. Among them, grid-based representation such as Plenoxels~\cite{fridovich2022plenoxels}, direct voxel grid optimization~\cite{sun2022direct} show superior performance in terms of training time, where it takes only a few minutes to learn plausible radiance fields. Combining grid representation with a hash function further improves the efficiency and training speed of the feature grids~\cite{mueller2022instant}. Fang \etal~\cite{fang2022fast} firstly applied a voxel grid representation to train dynamic NeRFs and achieved much faster training speed compared to the neural-network-based methods. Guo \etal~\cite{Guo_2022_NDVG_ACCV} also proposed a deformable voxel grid method for fast training. Unlike aforementioned works, our grid representation does not estimate warping or deformation. Instead, it directly estimates color and density using the features obtained from 4D hash grids, which decreases the computational burden and enables faster training.

\section{Temporal Interpolation for Dynamic NeRF}
\subsection{Preliminaries}
\label{subsec:pre}

Given 3D position $\mathbf{x} \in \mathbb{R}^3$ and 2D viewing direction $\mathbf{d} \in \mathbb{R}^2$, NeRF~\cite{mildenhall2021nerf} aims to estimate volume density $\sigma \in \mathbb{R}$ and emitted RGB color $\mathbf{c} \in \mathbb{R}^3$ using neural networks. We formulate dynamic NeRF as a direct extension from 3D to 4D by adding time frame $t \in \mathbb{R}$ to inputs, \ie
\begin{equation}
  (\mathbf{c}, \sigma) = \mathit{f}(\mathbf{v}(\mathbf{x}, t), \mathbf{d}).
\end{equation}
$\mathit{f}$ is implemented as fully connected neural networks and a space-time feature representation $\mathbf{v}$ can be MLP-based neural nets or explicit grid values as explained in \cref{subsec:stfr}.

From the camera with origin $\mathbf{o}$ and ray direction $\mathbf{d}$, the color of camera ray $\mathbf{r}(u) = \mathbf{o} + u\mathbf{d}$ at time frame $t$ is 
\begin{equation}
  \mathbf{C}(\mathbf{r},t) = \int_{u_n}^{u_f} U(\mathit{u},t)\sigma(\mathbf{v}) \mathbf{c}(\mathbf{v},\mathbf{d}) du,
\end{equation}
where $u_n$ and $u_f$ denote the bounds of the scene volume, $\mathbf{v}$ is an abbreviation of $\mathbf{v}(\mathbf{r}(u), t)$ and $U(\mathit{u},t)=\mathrm{exp}(-\int_{u_n}^{u}\sigma(\mathbf{v}(\mathbf{r}(s), t))ds)$ is the accumulated transmittance from ${u_n}$ to $u$.
Then, the RGB color loss $\mathit{L}_{c}$ is defined to minimize the $l_2$-loss between the estimated colors $\hat{C}(\mathbf{r}, t)$ and the ground truth colors $C(\mathbf{r}, t)$ over all rays $\mathbf{r}$ in camera views $R$ and time frames $t \in T$ as follows: 
\begin{equation}
    \mathit{L}_{c} = \sum_{\mathbf{r}\in R, t \in T} || \hat{C}(\mathbf{r}, t) - C (\mathbf{r}, t) ||^2_2.
\end{equation}

\subsection{Space-Time Feature Representation}
\label{subsec:stfr}
The main contribution of our paper lies on the novel feature representation method for dynamic NeRFs. We define the feature vector $\mathbf{v}$, which is fed into the neural network to determine a color and a volume density, as the concatenation of static and dynamic feature vectors, \ie,
\begin{equation}
    \mathbf{v}(\mathbf{x}, t) = [\mathbf{v}_s (\mathbf{x}), \mathbf{v}_d(\mathbf{x}, t)],
\end{equation}
where $[\cdot,\cdot]$ means concatenation operator of two vectors. The static feature vector $\mathbf{v}_s$ only depends on the 3D position $\mathbf{x}$. As most dynamic scenes also contain static regions such as background, $\mathbf{v}_s$ is designed to learn the static features that are consistent across time. Meanwhile, $\mathbf{v}_d$ learns dynamic features which may vary across time. We propose two novel feature representation methods in which temporal interpolation is applied. First, the neural representation, which is essentially a series of neural networks combined with linear interpolation, is suggested in \cref{subsec:nn}. The model trained using the neural representation is able to render high-quality space-time view synthesis with small-sized neural networks ($\sim$20MB). Second, the grid representation, which is an temporal extension of recently proposed voxel grid representations~\cite{mueller2022instant}, is explained in \cref{subsec:grid}. Dynamic NeRFs can be trained in less than 5 minutes with the proposed grid representation.

\subsubsection{Neural Representation}
\label{subsec:nn}

\begin{figure*}[t]
  \centering
   \includegraphics[width=0.95\linewidth]{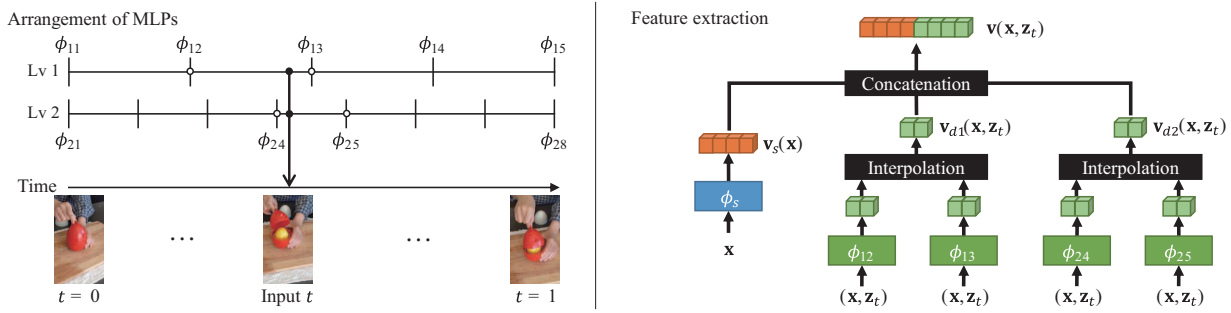}

   \caption{Illustration of the arrangements of MLPs (left) and the feature extraction (right) in the neural representation. $\mathit{\phi}_s$ and $\mathit{\phi}_{ij}$ are small MLPs that extract static and dynamic features respectively. The static features $\mathbf{v}_s$ and the dynamic feature $\mathbf{v}_d$ are concatenated and fed into the template NeRF. Detailed structures of the template NeRF can be found in the supplementary materials.}
   \label{fig:nn}
\end{figure*}

In the neural representation, features that fed into the template NeRFs are determined by a series of neural nets. In other words, both static and dynamic feature extractor are formulated as multi-layer perceptrons (MLPs). First, the whole time frame is divided into equally spaced time slots. For each time slot, two MLPs are assigned. Then the whole feature vector is interpolated from the assigned MLPs. Overall feature extraction procedure is illustrated in~\cref{fig:nn}.

Concretely, let $\mathbf{v}_d (\mathbf{x}, \mathbf{z}_t)$ be the feature vector for a 3D point $\mathbf{x}$ at time $t$, which will be fed to the template NeRF. Here, we used $\mathbf{z}_t$ as an embedding vector for an input time frame $\mathit{t}$~\cite{martin2021nerf,park2021hypernerf,li2022neural}. For equally spaced $n$ time slots, there are $n+1$ keyframes $t_i = \frac{i}{n}(i=0,1,2, \dots, n)$. An MLP $\phi_i$ is assigned to each keyframe $t_i$ which is responsible for two adjacent time slots $[t_{i-1},t_i]$ and $[t_i,t_{i+1}]$. For inputs with time $t \in [t_i,t_{i+1}]$, $\mathbf{x}$ and $\mathbf{z}_t$ are fed into $\phi_i$ and $\phi_{i+1}$. Then, the outputs from two MLPs are interpolated as follows:
\begin{equation}
    \mathbf{v}_d (\mathbf{x}, \mathbf{z}_t) = \Delta t \cdot \phi_{i}(\mathbf{x}, \mathbf{z}_t) + (1 - \Delta t) \cdot \phi_{i+1}(\mathbf{x}, \mathbf{z}_t)
\end{equation}
where $\Delta t = (t_{i+1}-t)/(t_{i+1}-t_{i})$.

The purpose of this interpolation is to efficiently learn the features between keyframes in a scalable manner. Thanks to the temporal interpolation, we can learn the features of the continuous time range $[t_i, t_{i+1}]$ by enforcing the MLPs $\phi_i$ and $\phi_{i+1}$ responsible for that time range. While the static feature $\mathbf{v}_s$ represents the features across the whole timeline, $\mathbf{v}_d$ represents the features that are focused more on dynamic regions. In addition, since each MLP for dynamic feature is responsible only for two adjacent time slots, it is able to make each $\phi_i$ learn features that are specific to a certain period of time.


To exploit the features of both long-term and short-term, there can be multiple levels of dynamic features which have different number of keyframes. For multi-level dynamic feature extractor with level $l$, each level contains different number of keyframes $n_1, n_2, \cdots, n_l$. Let $\mathbf{v}_{di}$ denote the dynamic feature of $i$-th level, then the output dynamic feature is the concatenation of features from all levels, \ie,
\begin{equation}
    \mathbf{v}_d (\mathbf{x}, \mathbf{z}_t) = [\mathbf{v}_{d1} (\mathbf{x}, \mathbf{z}_t), \mathbf{v}_{d2} (\mathbf{x}, \mathbf{z}_t), \cdots, \mathbf{v}_{dl} (\mathbf{x}, \mathbf{z}_t)] .
\end{equation}

In this paper, we used the settings $l=2, n_1=5, n_2=20$ otherwise stated. The dimensions of feature vectors are determined as 128 for $\mathbf{v}_s$, and 64 for $\mathbf{v}_0$ and $\mathbf{v}_1$. MLP with one hidden layer whose hidden size is the same as its output dimension is used for feature extraction of both $\mathbf{v}_s$ and $\mathbf{v}_{di}$. The concatenated feature vectors are fed to the template NeRF which outputs volume density $\sigma$ and emitted color $\mathbf{c}$. We used the same structure as used in HyperNeRF~\cite{park2021hypernerf} for the template NeRF of the neural representation.

\begin{figure*}[t]
  \centering
   \includegraphics[width=0.9\linewidth]{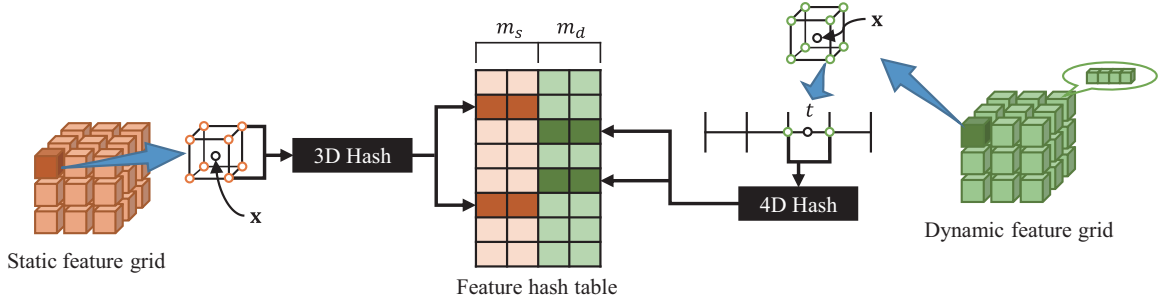}

   \caption{Illustration of grid representation with hash tables at a certain level. a static feature vector of a 3D grid point is extracted via 3D hash function, while dynamic feature vector for 4D grid point is extracted via 4D hash. The static feature vector $\mathbf{v}_s$ is determined as trilinear interpolation of 8 vectors of grid points, while the dynamic feature vector $\mathbf{v}_d$ is calculated as quadrilinear interpolation of 16 vectors. In the figure, we only depicted feature retrieval process of two grid points for illustration purpose.}
   \label{fig:grid}
\end{figure*}

\subsubsection{Grid Representation}
\label{subsec:grid}

Recently, InstantNGP~\cite{mueller2022instant} suggested a novel multi-level grid representation with hash tables. We adopt the hash grid representation from~\cite{mueller2022instant} and extend it for fast dynamic NeRF training. Similar to the neural representation in \cref{subsec:nn}, the feature vector from the proposed hash grid contains static and dynamic feature vectors. The static feature $\mathbf{v}_s$ is analogous to the one used in~\cite{mueller2022instant}. On the other hand, the dynamic feature $\mathbf{v}_d$ comes from the 4D hash grids.

Concretely, to extract static and dynamic feature vectors whose dimensions are $m_s$ and $m_d$ respectively, a hash table of size $H$ that contains $(m_s + m_d)$ dimension feature vectors is constructed. The hash function of $d$-dimensional vector, $h_d (\mathbf{x})$, is defined as
\begin{equation}
    \mathit{h}_{d} (\mathbf{x}) = \bigoplus_{i=1}^{d}( x_i P_i ) \quad \mathrm{mod} \,\, H,
\end{equation}
where $P_i$ is a large prime number and $\bigoplus$ is an XOR operator. Then, the feature vector $\mathbf{v}$ is retrieved by concatenating the outputs of 3D and 4D hash functions:
\begin{equation}
    \mathbf{v} (\mathbf{x},t) = [\mathit{h}_{3} (\mathbf{x}), \mathit{h}_{4} (\mathbf{x},t)].
\end{equation}

The 3D and 4D grids are constructed as the multi-level hash grids proposed in~\cite{mueller2022instant}. We applied different scaling factors for 3D space and time frame since the number of frames for training sequences are usually much smaller than the finest 3D grid resolutions.

\subsection{Smoothness Regularization}

As our dynamic world smoothly changes, it is reasonable to impose a smoothness term to adjacent time frames. We note that the smoothness term is only applied to the input feature space, not to the estimated outputs such as RGB color or density. For the neural representation, we provided a simple smoothness term which is calculated as,
\begin{equation}
    \mathit{L}_{s} (\mathbf{x}, t) = \lVert \mathbf{v}_d (\mathbf{x}, \mathbf{z}_t) - \mathbf{v}_d (\mathbf{x}, \mathbf{z}_{t+1}) \rVert _2 ^2 .
    \label{eq:smooth_nn}
\end{equation}
This regularization term has two advantages. First, it reflects the intuition that the observed point $\mathbf{x}$ at time $t$ will be stationary if there is no observation for $\mathbf{x}$ at time $t+1$. Second, by propagating additional gradients to the feature networks or the hash grids, the smoothness term acts as a regularizer that stabilizes the training.

For the grid representation, we impose a smoothness term to the grid points that are temporally adjacent:
\begin{equation}
    \mathit{L}_{s} (\mathbf{x}, t) = \frac{1}{n_f ^ 2} \lVert h_4 (\mathbf{x}, t_a ) - h_4 (\mathbf{x}, t_b ) \rVert _2 ^2 ,
    \label{eq:smooth_grid}
\end{equation}
where $n_f$ is the number of frames in the training sequence and $t_a , t_b$ are two adjacent grid points that satisfies $t_a \leq t \leq t_b$. In fact, \cref{eq:smooth_grid} can be obtained from \cref{eq:smooth_nn} in the grid representation, which indicates that imposing smoothness term to adjacent time frames is equivalent to add smoothness to two temporally adjacent grids with constant multiplier. Detailed derivation of this relationship is provided in the supplementary materials.

The smoothness term is added to the loss function with a weight $\lambda$, and the total loss is minimized together during the training. 
Accordingly, the total loss of our dynamic neural radiance fields is given by
\begin{equation}
    \mathit{L} = \mathit{L}_{c} + \lambda \mathit{L}_{s} .
\end{equation}

\begin{figure}
  \centering
  \hfill
  \begin{subfigure}{0.31\linewidth}
     \includegraphics[width=0.95\linewidth]{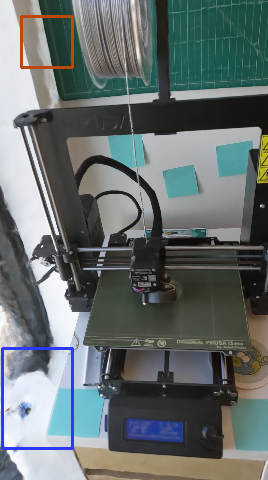}
    \caption{w/o smooth term}
    \label{fig:short-a}
  \end{subfigure}
  \hfill
  \begin{subfigure}{0.31\linewidth}
     \includegraphics[width=0.95\linewidth]{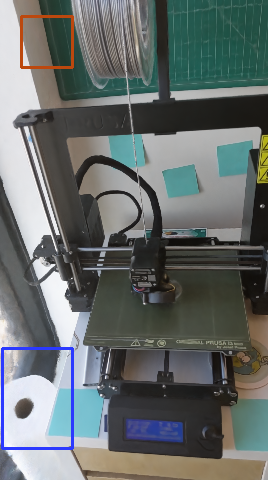}
    \caption{w/ smooth term}
    \label{fig:short-b}
  \end{subfigure}
  \hfill
  \begin{subfigure}{0.31\linewidth}
     \includegraphics[width=0.95\linewidth]{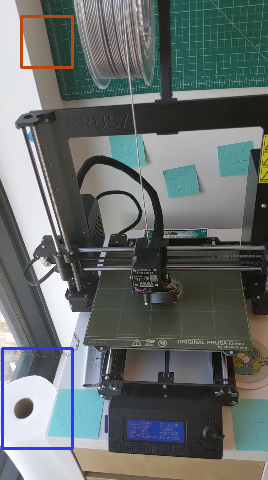}
    \caption{GT}
    \label{fig:short-c}
  \end{subfigure}
  \hfill
  \caption{Effectiveness of the smoothness term. It can be observed that the model trained with the smoothness term accurately renders the corner part which appears in only few frames within the whole sequence (blue box) and removes spurious artifacts (red box).}
  \vspace{-0.5em}
  \label{fig:smooth}
\end{figure}

We observe that the smoothness term is especially powerful in the static background regions that appeared only in a small fraction of frames. In~\cref{fig:smooth}, we show examples of the rendered images which are trained with and without the smoothness term. It can be clearly seen that the boxed regions of the rendered image shows much plausible when the model is trained with the smoothness term.

\subsection{Implementation Details}
In the neural representation, we adopted the template NeRF architecture of HyperNeRF~\cite{park2021hypernerf} for density and color estimation. The network is an 8-layered MLP with hidden size of 256. The dimension of the embedding vector $\mathbf{z}_t$ is set to 8. The smoothness weight $\lambda$ is set to 0.01 or 0.001 depending on the characteristics of the datasets used in the experiments. We set $\lambda$ to a large value in the sequences that the viewpoint change is significant in a short period of time although performance variations depending on the value of $\lambda$ is not significant.

In the grid representation, $\lambda$ is set to $1e^{-4}$. We applied the smoothness loss only for the finest two levels of the temporal dimension since applying it to every level slows training speed without much improvements in performance. After feature vectors are extracted from the hash grids, we fed them to a 3-layered MLP with hidden size of 128 to estimate volume density followed by one additional layer for RGB color estimation similar to~\cite{mueller2022instant}. We set $m_s = 2, m_d = 6$ and $H=2^{19}$ for the  grid representation experiments otherwise stated. The 3D and 4D hash grids are composed of 12 levels. For the spatial dimension, we set the base resolution to 8, and the scaling factor is set to 1.45. The base resolution of the temporal dimension is 2, which is multiplied by 1.4 in every other level. Detailed hyperparameter settings and network architectures can be found in the supplementary materials.

\section{Experimental Results}
\subsection{Datasets}

To validate the superiority of our method, we conducted extensive experiments on various datasets. We used three publicly available datasets that are used for dynamic NeRFs. For all experiments, we trained our models for each sequence individually, and then per-scene results are averaged and reported in this section.

{\bf \noindent D-NeRF Dataset~\cite{pumarola2021d}.} The dataset consists of synthetically rendered images of moving and deforming 3D objects. For each frame, the image is rendered via a synthetic camera of random rotation. There are eight scenes in total, and each scene contains 50-200 training images with 20 test views.

{\bf \noindent HyperNeRF Dataset~\cite{park2021hypernerf}.} The dataset contains video sequences taken from mobile phones. There are four sequences in \textit{vrig-dataset} which are taken using the camera rig with two phones vertically aligned. In addition, there are six sequences in \textit{interp-dataset} which are taken from a single camera in order to estimate the image in the the middle of two consecutive training images.

{\bf \noindent DyNeRF Dataset~\cite{li2022neural}.} The dataset consists of videos obtained from a capture system that consists of 21 GoPro Black Hero cameras. The cameras are located at a fixed position, and all video frames are synchronized to build the multi-view video dataset.

\subsection{Neural Representation}

In this section, we reported the performance of the proposed neural representation models. First, the experimental results on D-NeRF dataset are shown in~\cref{tab:dnerf}. Peak signal-to-noise ratio (PSNR), structural similarity (SSIM), and perceptual similarity (LPIPS)~\cite{zhang2018perceptual} are used as evaluation metrics following the previous works. We also reported the average metric (AVG) proposed in~\cite{barron2021mip} which aggregates three metrics to a single value. Our method with the neural representation (Ours-NN) achieves state-of-the-art results on all evaluation metrics. It is worth noting that the smoothness term dramatically improves overall performance.

Next, we evaluated our method on HyperNeRF dataset~\cite{park2021hypernerf} and compared with existing methods, which is shown in~\cref{tab:hypernerf}. Here, we reported PSNR and multi-scale SSIM (MS-SSIM), and excluded LPIPS metric since its value cannot reliably reproduced~\cite{fang2022fast}. Our method shows second-best results on both \textit{vrig} and \textit{interp} datasets. While flow-based method~\cite{li2021neural} suffers from interpolating motions between two consecutive frames, our method, which implicitly learns intermediate feature representation of in-between frames, achieves competitive performance with warping-based methods~\cite{park2021nerfies,park2021hypernerf}.

\begin{figure*}
  \centering
  \begin{subfigure}{0.12\linewidth}
     \includegraphics[width=0.97\linewidth]{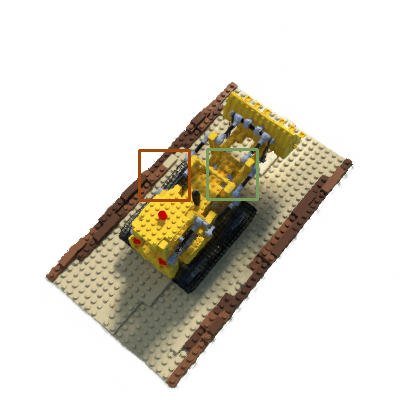}
  \end{subfigure}
  \begin{subfigure}{0.05\linewidth}
     \includegraphics[width=0.95\linewidth]{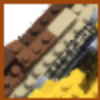}
     \includegraphics[width=0.95\linewidth]{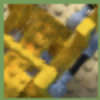}
  \end{subfigure}
  \hfill
  \begin{subfigure}{0.12\linewidth}
     \includegraphics[width=0.97\linewidth]{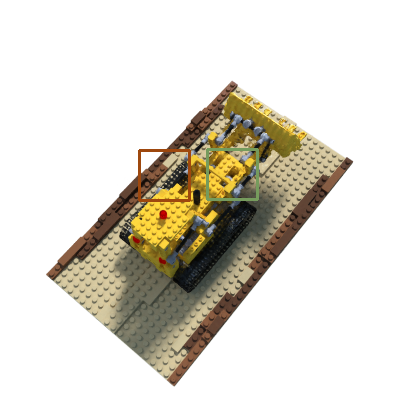}
  \end{subfigure}
  \begin{subfigure}{0.05\linewidth}
     \includegraphics[width=0.95\linewidth]{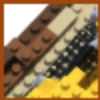}
     \includegraphics[width=0.95\linewidth]{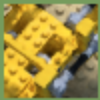}
  \end{subfigure}
  \hfill
  \begin{subfigure}{0.12\linewidth}
     \includegraphics[width=0.97\linewidth]{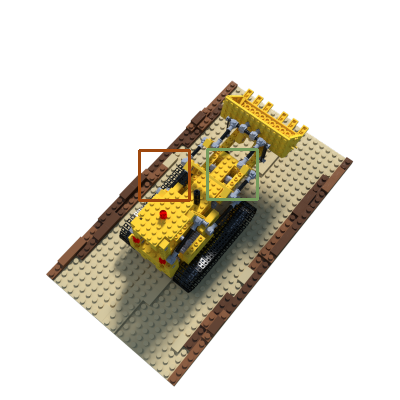}
  \end{subfigure}
  \begin{subfigure}{0.05\linewidth}
     \includegraphics[width=0.95\linewidth]{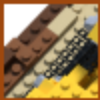}
     \includegraphics[width=0.95\linewidth]{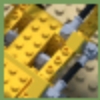}
  \end{subfigure}
  \hfill
  \begin{subfigure}{0.08\linewidth}
     \includegraphics[width=0.95\linewidth]{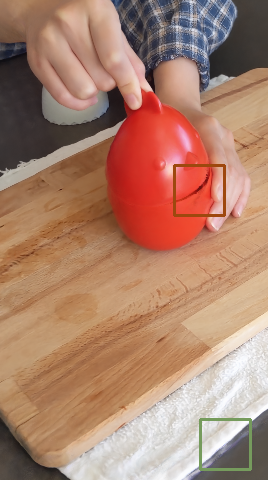}
  \end{subfigure}
  \begin{subfigure}{0.05\linewidth}
     \includegraphics[width=0.95\linewidth]{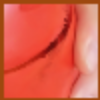}
     \includegraphics[width=0.95\linewidth]{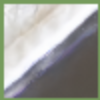}
  \end{subfigure}
  \hfill
  \begin{subfigure}{0.08\linewidth}
     \includegraphics[width=0.95\linewidth]{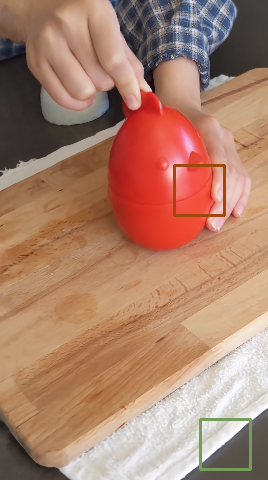}
  \end{subfigure}
  \begin{subfigure}{0.05\linewidth}
     \includegraphics[width=0.95\linewidth]{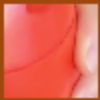}
     \includegraphics[width=0.95\linewidth]{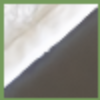}
  \end{subfigure}
  \hfill
  \begin{subfigure}{0.08\linewidth}
     \includegraphics[width=0.95\linewidth]{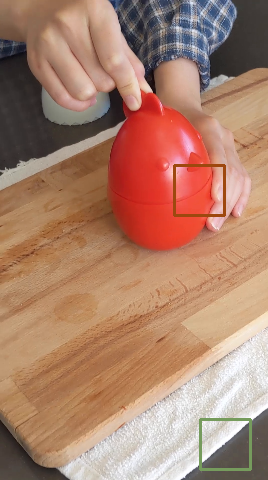}
  \end{subfigure}
  \begin{subfigure}{0.05\linewidth}
     \includegraphics[width=0.95\linewidth]{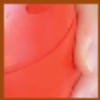}
     \includegraphics[width=0.95\linewidth]{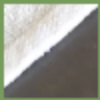}
  \end{subfigure}

  \begin{subfigure}{0.12\linewidth}
     \includegraphics[width=0.97\linewidth]{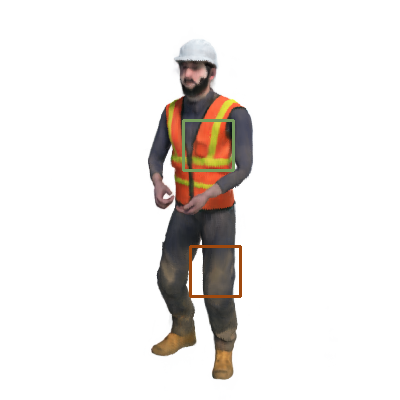}
  \end{subfigure}
  \begin{subfigure}{0.05\linewidth}
     \includegraphics[width=0.95\linewidth]{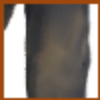}
     \includegraphics[width=0.95\linewidth]{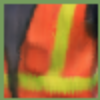}
  \end{subfigure}
  \hfill
  \begin{subfigure}{0.12\linewidth}
     \includegraphics[width=0.97\linewidth]{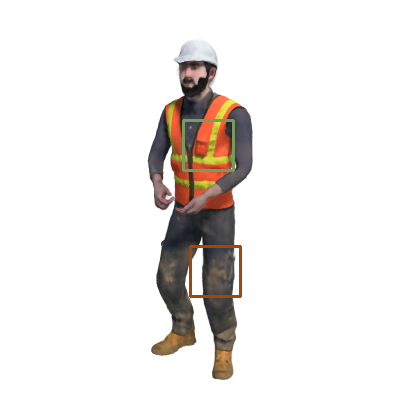}
  \end{subfigure}
  \begin{subfigure}{0.05\linewidth}
     \includegraphics[width=0.95\linewidth]{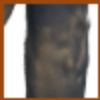}
     \includegraphics[width=0.95\linewidth]{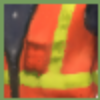}
  \end{subfigure}
  \hfill
  \begin{subfigure}{0.12\linewidth}
     \includegraphics[width=0.97\linewidth]{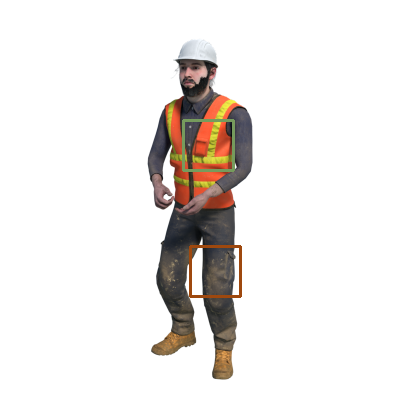}
  \end{subfigure}
  \begin{subfigure}{0.05\linewidth}
     \includegraphics[width=0.95\linewidth]{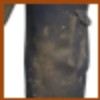}
     \includegraphics[width=0.95\linewidth]{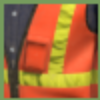}
  \end{subfigure}
  \hfill
  \begin{subfigure}{0.08\linewidth}
     \includegraphics[width=0.95\linewidth]{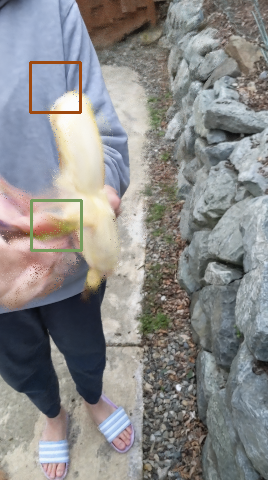}
  \end{subfigure}
  \begin{subfigure}{0.05\linewidth}
     \includegraphics[width=0.95\linewidth]{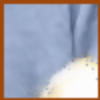}
     \includegraphics[width=0.95\linewidth]{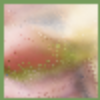}
  \end{subfigure}
  \hfill
  \begin{subfigure}{0.08\linewidth}
     \includegraphics[width=0.95\linewidth]{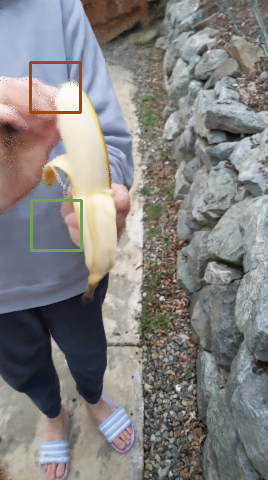}
  \end{subfigure}
  \begin{subfigure}{0.05\linewidth}
     \includegraphics[width=0.95\linewidth]{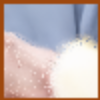}
     \includegraphics[width=0.95\linewidth]{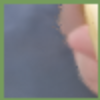}
  \end{subfigure}
  \hfill
  \begin{subfigure}{0.08\linewidth}
     \includegraphics[width=0.95\linewidth]{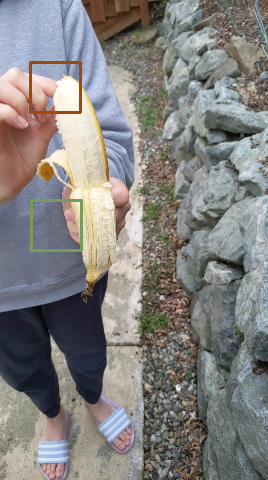}
  \end{subfigure}
  \begin{subfigure}{0.05\linewidth}
     \includegraphics[width=0.95\linewidth]{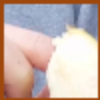}
     \includegraphics[width=0.95\linewidth]{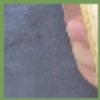}
  \end{subfigure}
  
  \small
  \begin{subfigure}{0.17\linewidth}
  \quad \quad \quad D-NeRF~\cite{pumarola2021d}
  \end{subfigure}
  \hfill
  \begin{subfigure}{0.17\linewidth}
  \quad \quad \quad \quad Ours-NN
  \end{subfigure}
  \hfill
  \begin{subfigure}{0.17\linewidth}
  \quad \quad \quad \quad Ground truth
  \end{subfigure}
  \hfill
  \begin{subfigure}{0.16\linewidth}
  \quad \, \, HyperNeRF~\cite{park2021hypernerf}
  \end{subfigure}
  \hfill
  \begin{subfigure}{0.14\linewidth}
  \quad \quad \, Ours-NN
  \end{subfigure}
  \hfill
  \begin{subfigure}{0.14\linewidth}
  \quad Ground truth
  \end{subfigure}
  
  \caption{Qualitative results of our method (Ours-NN) on of D-NeRF~\cite{pumarola2021d} and HyperNeRF~\cite{park2021hypernerf} datasets. The regions that show significant difference with compared methods are scaled up (red and green boxes).}
  \label{fig:qual_nn}
\end{figure*}

Qualitative results for the neural representation are shown in~\cref{fig:qual_nn}. It is clearly observed that our method captures fine details compared to D-NeRF~\cite{pumarola2021d}. We highlighted the regions that show notable differences with colored boxes and placed zoomed-in images of the regions next to the rendered images. In addition, we also qualitatively compared our method with HyperNeRF~\cite{park2021hypernerf} on their datasets. When the warping estimation module of \cite{park2021hypernerf} does not correctly estimate warping parameters, HyperNeRF produces implausible results. It can be observed that the head and the body of the chicken toy is not properly interlocked and the position of the hand peeling banana is incorrect. On the other hand, our method accurately recovers 3D geometry of those scenes. Thus, without using the physically meaningful warping estimation module in neural networks, the proposed temporal interpolation and the smoothness regularizer provides simple yet effective way to learn complex deformations.

Lastly, performance evaluation on DyNeRF dataset~\cite{li2022neural} is presented in~\cref{tab:dynerf}. We adopt PSNR, LPIPS, and  FLIP~\cite{Andersson2020} for evaluation metrics to compare with previous works. Our method achieves best PSNR while ranked second in LPIPS and FLIP. However, those two metrics are also better than DyNeRF$^\dagger$~\cite{li2022neural} which does not use importance sampling strategy in~\cite{li2022neural}. Since we do not use any sampling strategy during training, it can be concluded that our feature interpolation method is superior to the network architecture of~\cite{li2022neural}. Notably, our method outperforms DyNeRF with smaller network size (20MB) than the DyNeRF models (28MB).

\begin{table}[t]
\centering
\resizebox{0.47\textwidth}{!}{
    \begin{tabular}{l|c|c|c|c}
     \quad & PSNR$\,\uparrow$        & SSIM$\,\uparrow$         & LPIPS$\,\downarrow$        & AVG$\,\downarrow$           \\ \hline
    NeRF~\cite{mildenhall2021nerf} & 19.00          & 0.87          & 0.18          & 0.09          \\
    T-NeRF~\cite{pumarola2021d}  & 29.50          & 0.95          & 0.08          & 0.03          \\
    D-NeRF~\cite{pumarola2021d}  & 30.43          & 0.95          & 0.07          & 0.02          \\
    NDVG-full~\cite{Guo_2022_NDVG_ACCV}        & 27.84          & 0.862 & 0.041          & 0.029          \\
    TiNeuVox-B~\cite{fang2022fast}        & {\ul 32.67}          & {\ul 0.971} & 0.041          & {\ul 0.016}          \\ \hline
    Ours-NN (w/o smooth)                   & 30.18 & 0.963 & {\ul 0.038} & 0.019 \\
    Ours-NN (w/ smooth)                    & \textbf{32.73} & \textbf{0.974} & \textbf{0.033} & \textbf{0.014}
    \end{tabular}
}
\caption{Experimental results on D-NeRF~\cite{pumarola2021d} datasets.}
\vspace{-0.5em}
\label{tab:dnerf}
\end{table}

\begin{table}[t]
\centering
\resizebox{0.47\textwidth}{!}{
    \begin{tabular}{l|c|c|c|c}
      \quad  & \multicolumn{2}{c}{\textit{vrig}}                              & \multicolumn{2}{c}{\textit{interp}}                            \\
       \quad & \multicolumn{1}{c}{PSNR$\,\uparrow$} & \multicolumn{1}{c}{MS-SSIM$\,\uparrow$} & \multicolumn{1}{c}{PSNR$\,\uparrow$} & \multicolumn{1}{c}{MS-SSIM$\,\uparrow$} \\ \hline
    NeRF~\cite{mildenhall2021nerf}       & 20.13                     & 0.745                     & 22.27                     & 0.804                     \\
    NV~\cite{Lombardi:2019}         & 16.85                     & 0.571                     & 26.05                     & 0.911                     \\
    NSFF~\cite{li2021neural}       & \textbf{26.33}            & \textbf{0.916}            & 25.80                     & 0.883                     \\
    Nerfies~\cite{park2021nerfies}    & 22.23                     & 0.803                     & 28.47                     & 0.939                     \\
    HyperNeRF~\cite{park2021hypernerf} & 22.38                     & 0.814                     & \textbf{29.00}            & \textbf{0.945}            \\ \hline
    Ours-NN & {\ul 24.35}               & {\ul 0.867}               & {\ul 28.67}               & {\ul 0.940}              
    \end{tabular}
}
\caption{Experimental results on HyperNeRF~\cite{park2021hypernerf} datasets.}
\label{tab:hypernerf}
\end{table}

\begin{table}[t]
\centering
\resizebox{0.42\textwidth}{!}{
    \begin{tabular}{l|c|c|c}
    \,        & {PSNR$\,\uparrow$}              & {LPIPS$\,\downarrow$}           & {FLIP$\,\downarrow$}            \\ \hline
    MVS           & 19.12          & 0.2599          & 0.2542          \\
    NeuralVolumes~\cite{Lombardi:2019} & 22.80           & 0.2951          & 0.2049          \\
    LLFF~\cite{mildenhall2019local}          & 23.24           & 0.2346          & 0.1867          \\
    NeRF-T~\cite{li2022neural}        & 28.45           & 0.100             & 0.1415          \\
    DyNeRF$^\dagger$~\cite{li2022neural}       & 28.50           & 0.0985          & 0.1455          \\
    DyNeRF~\cite{li2022neural}        & {\ul 29.58}           & \textbf{0.0832} & \textbf{0.1347} \\ \hline
    Ours-NN       & \textbf{29.88} & {\ul 0.0960}        & {\ul 0.1413}       
    \end{tabular}
}
\caption{Experimental results on DyNeRF~\cite{li2022neural} datasets.}
\vspace{-0.5em}
\label{tab:dynerf}
\end{table}

\subsection{Grid Representation}

We used D-NeRF datasets to evaluate the performance of the proposed grid representation. Since this representation is mainly intended for fast training, we report the results in a short period time ($\sim$8 minutes) and compare the results with the concurrent works~\cite{fang2022fast,Guo_2022_NDVG_ACCV}, both of which are based on voxel grid representation and showed the fastest training speed for dynamic NeRFs so far. We also examined the performance of the original implementation of InstantNGP~\cite{mueller2022instant} as a baseline with no temporal extension. All of the grid representation models in the experiments are trained on a single RTX 3090 GPU for fair comparison with~\cite{fang2022fast,Guo_2022_NDVG_ACCV}.

\begin{table}[t]
\centering
\resizebox{0.475\textwidth}{!}{
\begin{tabular}{l|c|c|c|c|c}
\,     & Train time & PSNR  & SSIM  & LPIPS & AVG   \\ \hline
InstantNGP~\cite{mueller2022instant} & 5 min         & 20.28 & 0.888 & 0.146 & 0.077 \\ \hline
D-NeRF~\cite{pumarola2021d} & 20 hours   & 30.43          & 0.95          & 0.07          & 0.02          \\
NDVG-full~\cite{Guo_2022_NDVG_ACCV}   & 35 min      & 27.84 &	0.862 &	0.041 &	0.029          \\
TinueVox-B~\cite{fang2022fast} & 30 min         & 32.67          & 0.971 & 0.041          & 0.016         \\
NDVG-half~\cite{Guo_2022_NDVG_ACCV}   & 23 min      & 27.15 &	0.857 &	0.048 &	0.033          \\
TinueVox-S~\cite{fang2022fast} & 8 min         & 30.75 & 0.956 & 0.067 & 0.023 \\ \hline
Ours-grid  & 1 min         & 26.77	&	0.933	&	0.107 &	0.039 \\
Ours-grid  & 5 min         & 29.73	&	0.961	&	0.063 &	0.024 \\
Ours-grid  & 8 min         & 29.84 & 0.962 & 0.062 & 0.023
\end{tabular}
}
\caption{Quantitative comparison of training time and performance on D-NeRF datasets using the grid representation.}
\vspace{-0.5em}
  \label{tab:grid}
\end{table}

\begin{figure*}
    \centering
      \begin{subfigure}{0.16\linewidth}
        \includegraphics[width=0.97\linewidth]{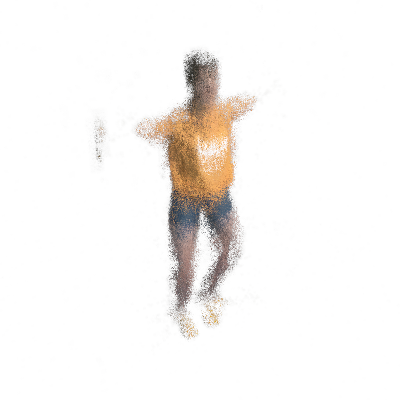}
      \end{subfigure}
      \hfill
      \begin{subfigure}{0.16\linewidth}
        \includegraphics[width=0.97\linewidth]{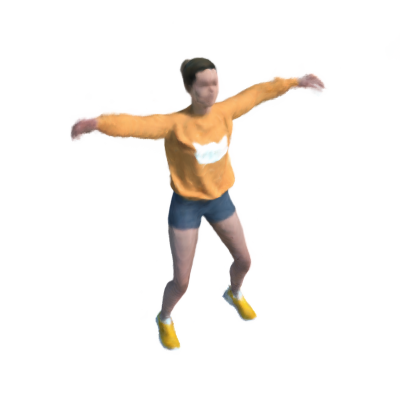}
      \end{subfigure}
  \unskip\ \vrule\ 
      \hfill
      \begin{subfigure}{0.16\linewidth}
        \includegraphics[width=0.97\linewidth]{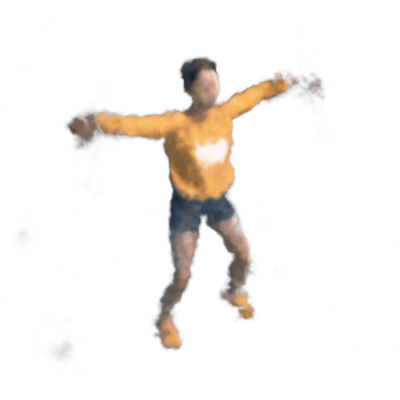}
      \end{subfigure}
      \hfill
      \begin{subfigure}{0.16\linewidth}
        \includegraphics[width=0.97\linewidth]{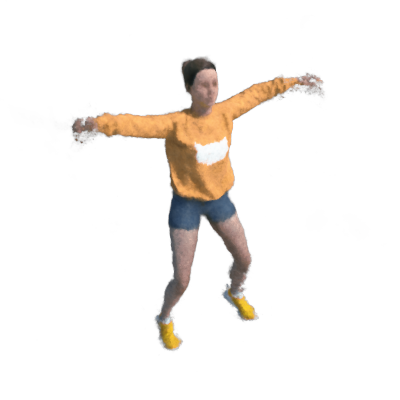}
      \end{subfigure}
  \vspace{-1.07\baselineskip}
      \hfill
      \begin{subfigure}{0.16\linewidth}
        \includegraphics[width=0.97\linewidth]{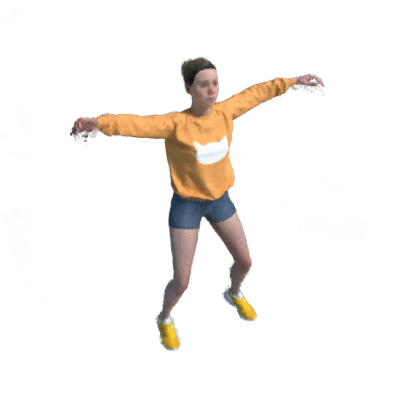}
      \end{subfigure}
  \unskip\ \vrule\ 
      \hfill
      \begin{subfigure}{0.16\linewidth}
        \includegraphics[width=0.97\linewidth]{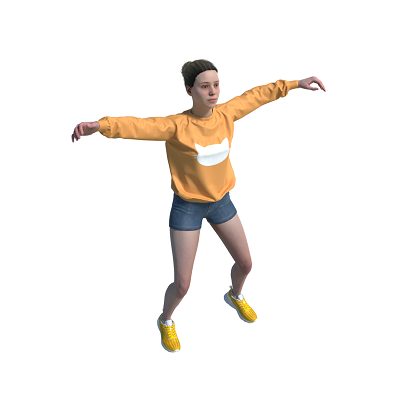}
      \end{subfigure}

    \centering
      \begin{subfigure}{0.16\linewidth}
        \includegraphics[width=0.97\linewidth]{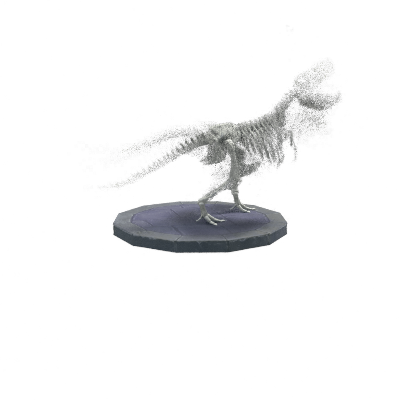}
      \end{subfigure}
      \hfill
      \begin{subfigure}{0.16\linewidth}
        \includegraphics[width=0.97\linewidth]{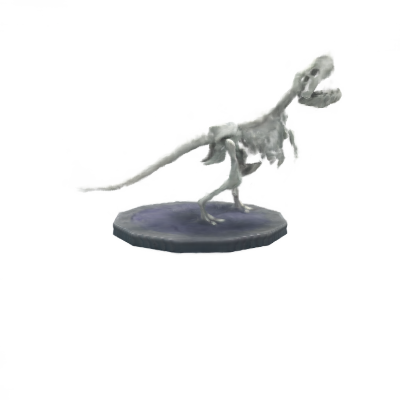}
      \end{subfigure}
  \unskip\ \vrule\ 
      \hfill
      \begin{subfigure}{0.16\linewidth}
        \includegraphics[width=0.97\linewidth]{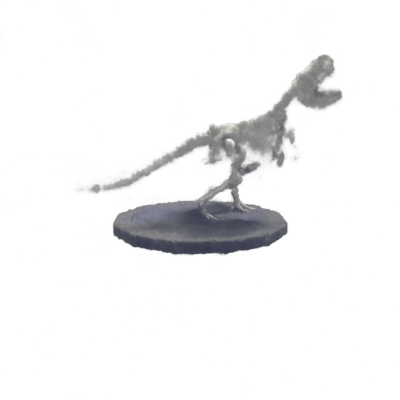}
      \end{subfigure}
      \hfill
      \begin{subfigure}{0.16\linewidth}
        \includegraphics[width=0.97\linewidth]{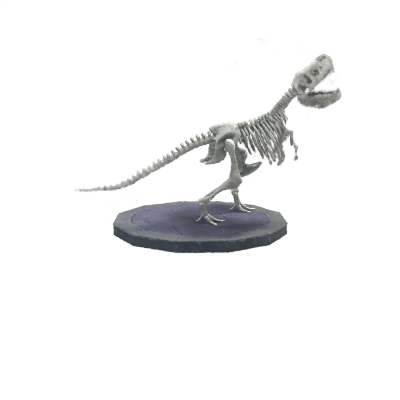}
      \end{subfigure}
  \vspace{-1.5\baselineskip}
      \hfill
      \begin{subfigure}{0.16\linewidth}
        \includegraphics[width=0.97\linewidth]{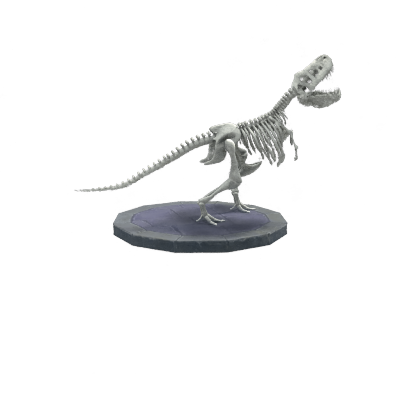}
      \end{subfigure}
  \unskip\ \vrule\ 
      \hfill
      \begin{subfigure}{0.16\linewidth}
        \includegraphics[width=0.97\linewidth]{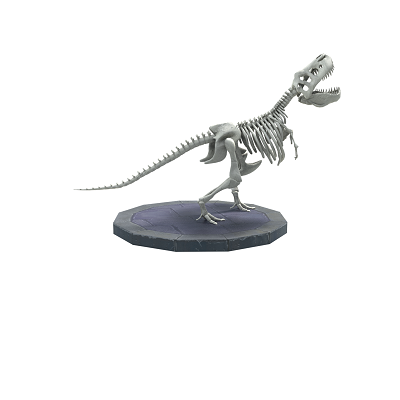}
      \end{subfigure}

  \small
  \begin{subfigure}{0.16\linewidth}
  \centering
  InstantNGP~\cite{mueller2022instant}
  \end{subfigure}
  \hfill
  \begin{subfigure}{0.16\linewidth}
  \centering
  TiNueVox-S~\cite{fang2022fast} (8min)
  \end{subfigure}
  \hfill
  \begin{subfigure}{0.16\linewidth}
  \centering
  Ours-grid (1min)
  \end{subfigure}
  \hfill
  \begin{subfigure}{0.16\linewidth}
  \centering
  Ours-grid (3min)
  \end{subfigure}
  \hfill
  \begin{subfigure}{0.16\linewidth}
  \centering
  Ours-grid (5min)
  \end{subfigure}
  \hfill
  \begin{subfigure}{0.16\linewidth}
  \centering
  Ground truth
  \end{subfigure}
  \hfill
  
    \caption{Qualitative results of the grid representation. While InstantNGP~\cite{mueller2022instant} is not able to reconstruct dynamic regions, our method produces acceptable rendering results in one minute and performs superior to TiNueVox-S~\cite{fang2022fast} with shorter training time.}
    \label{fig:qual_grid}
\end{figure*}

\begin{figure}
  \centering
  \begin{subfigure}{0.49\linewidth}
     \includegraphics[width=0.97\linewidth]{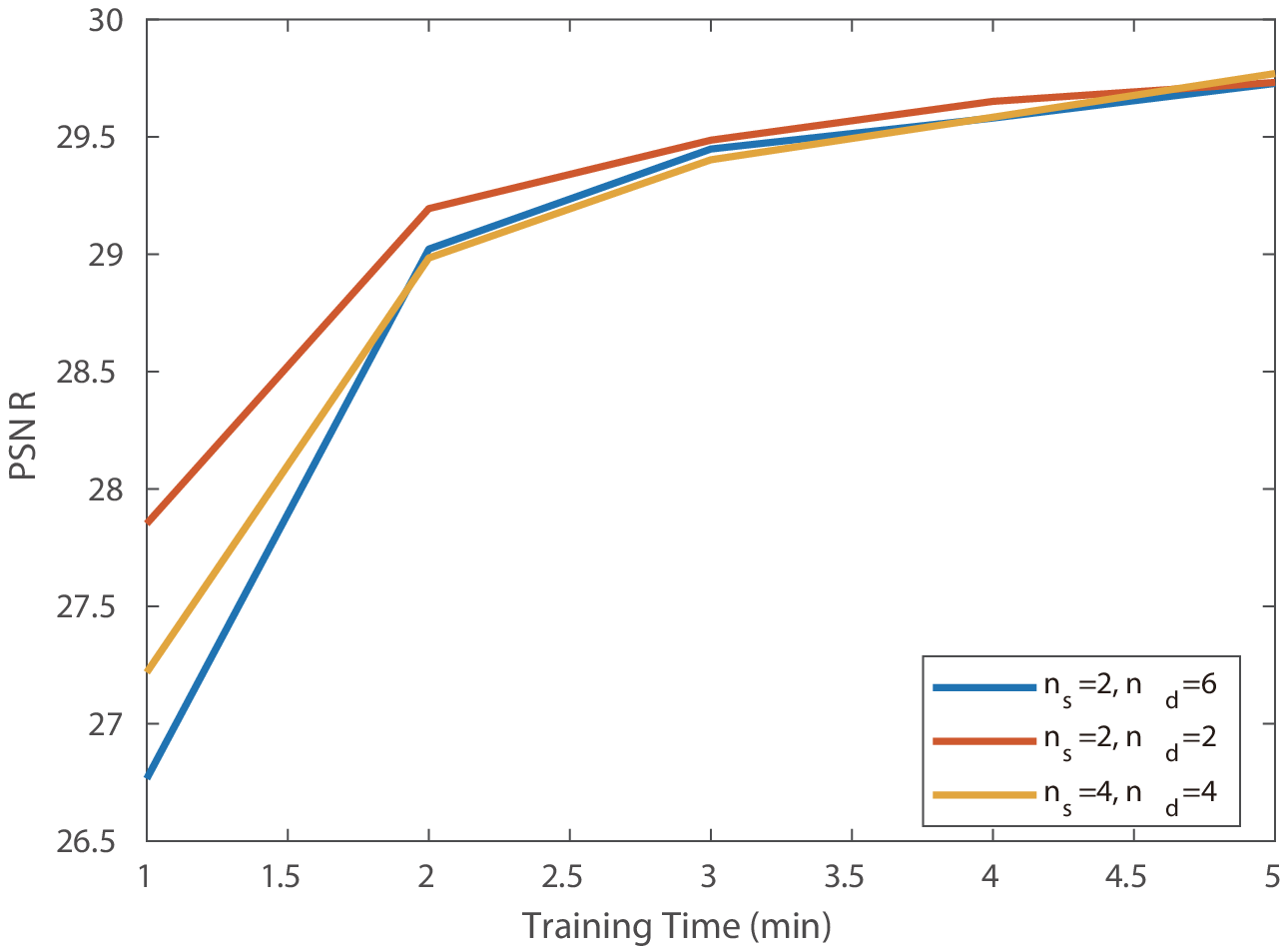}
  \end{subfigure}
  \hfill
  \begin{subfigure}{0.49\linewidth}
     \includegraphics[width=0.97\linewidth]{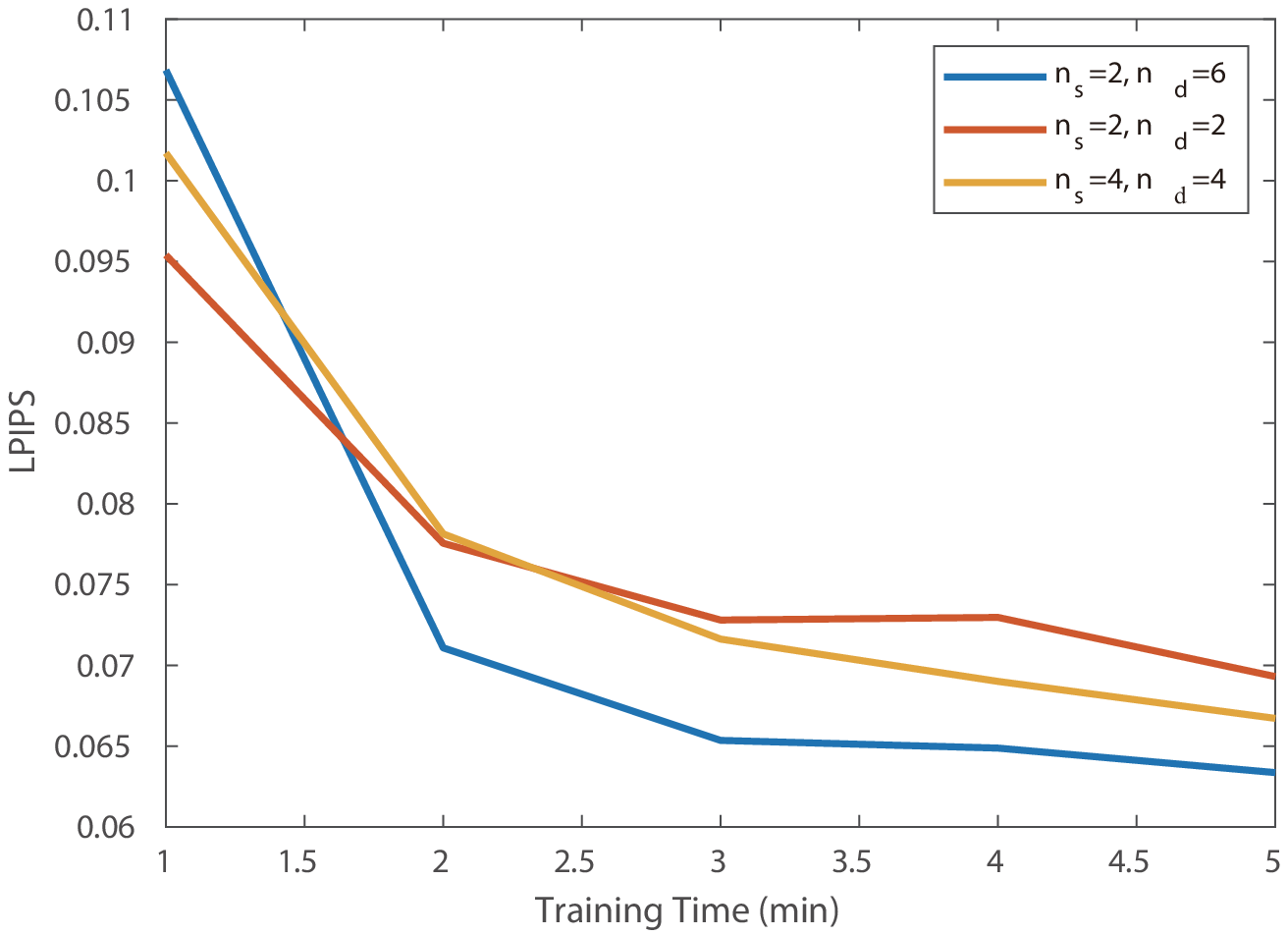}
  \end{subfigure}
    \caption{Performance comparison with different feature dimension settings. Using smaller feature dimensions results in faster training but quicker saturation of model performance.}
    \label{fig:chart}
    \vspace{-0.5em}
\end{figure}

\begin{table}[t]
\resizebox{0.47\textwidth}{!}{
    \begin{tabular}{l|c|c|c|c|c}
      \quad  & \multicolumn{3}{c}{D-NeRF dataset}                              & \multicolumn{2}{c}{HyperNeRF \textit{vrig}}                            \\
       \quad & \multicolumn{1}{c}{PSNR} & \multicolumn{1}{c}{SSIM} & \multicolumn{1}{c}{LPIPS} & \multicolumn{1}{c}{PSNR} & \multicolumn{1}{c}{MS-SSIM} \\ \hline
    NN (dynamic only)       & 30.91                     & 0.963        & 0.043             & 24.07                     & 0.861                     \\
    NN (dynamic+static)         & \bf 32.73                     & \bf 0.974       & \bf 0.033              & \bf 24.35                     & \bf 0.866         \\\hline
    Grid (dynamic only)      & 29.08                     & 0.948        & 0.076            &     22.11      &   0.755  \\
    Grid (dynamic+static)        & \bf 29.84                     & \bf 0.962       & \bf 0.062              &   \bf 22.98    &  \bf 0.802                     \\\hline
    \end{tabular}
}
\caption{Ablation study on the effectiveness of static features.}
\vspace{-0.5em}
\label{tab:exp_static}
\end{table}

The quantitative results are shown with elapsed training time in \cref{tab:grid}. The grid representation demonstrates much faster training speed even compared to the recent voxel-grid based methods~\cite{fang2022fast,Guo_2022_NDVG_ACCV}. In comparison with neural network models~\cite{pumarola2021d}, our method exceeds SSIM and LPIPS of~\cite{pumarola2021d} in 5 minutes which have arithmetically 240 times faster training speed. By taking the benefits of fully fused neural network~\cite{mueller2021real} as well as the efficient hash grid representation, our method quickly learns meaningful features in dynamic scenes and generates acceptable novel view synthesis results with just one minute of training. Our method shows superior SSIM, LPIPS and slightly inferior PSNR to TinueVox-S~\cite{fang2022fast}, which indicates clearer and more detailed rendering results, when trained for the same period of time.

We depict training progress of the grid representation qualitatively in~\cref{fig:qual_grid}. We also compared the results of TinueVox-S~\cite{fang2022fast} which are trained for 8 minutes. After one minute of training, the grid representation model produces blurry images but accurately render dynamic regions compared to InstantNGP~\cite{mueller2022instant}. After 5 minutes, the model successfully renders sharp images that are similar to the ground truth images. Notably, our model exhibits more sharp and clear results than TinueVox-S despite shorter training time, which results in better SSIM and LPIPS..

Finally, we compared the performance of the grid representation by varying the dimension of static and dynamic feature vectors, $m_s$ and $m_d$. PSNR and LPIPS on test images are measured per minute and illustrated in \cref{fig:chart}. We trained the grid representation models in three different settings, $m_s=2, m_d=6$, $m_s=2, m_d=2$, $m_s=4, m_d=4$. When smaller feature dimension is used, the training speed is faster, so PSNR increases fast in the early stage of training. However, the performance also saturates faster, and LPIPS is inferior to the other settings. The model with $m_s=2, m_d=6$ significantly outperformed the others in terms of LPIPS. 

\subsection{Effectiveness of the Static Features}
To validate the effectiveness of the static features, we conducted an ablation study on the static features and illustrated the results in \cref{tab:exp_static}. For both the neural representation (NN) and the grid representation (Grid), the models with static features performs superior to the ones using only dynamic features in all metrics. Additional ablation studies and qualitative results can be found in the supplementary materials.

\subsection{Failure Cases}

\begin{figure}
    \centering
      \begin{subfigure}{0.23\linewidth}
     \includegraphics[width=0.97\linewidth]{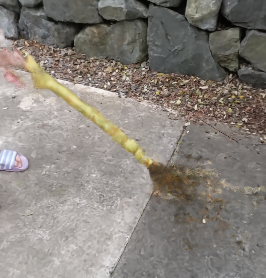}
  \end{subfigure}
      \begin{subfigure}{0.23\linewidth}
     \includegraphics[width=0.97\linewidth]{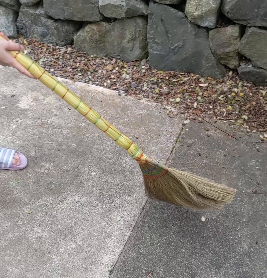}
  \end{subfigure}
 \hfill
      \begin{subfigure}{0.23\linewidth}
     \includegraphics[width=0.97\linewidth]{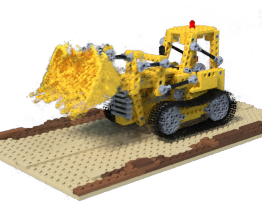}
  \end{subfigure}
      \begin{subfigure}{0.23\linewidth}
     \includegraphics[width=0.97\linewidth]{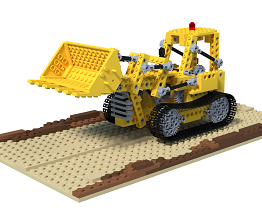}
  \end{subfigure}
  
  \small
  \begin{subfigure}{0.23\linewidth}
  \centering
  Ours-NN
  \end{subfigure}
  \begin{subfigure}{0.23\linewidth}
  \centering
  Ground truth
  \end{subfigure}
  \hfill
  \begin{subfigure}{0.23\linewidth}
  \centering
  Ours-grid
  \end{subfigure}
  \begin{subfigure}{0.23\linewidth}
  \centering
  Ground truth
  \end{subfigure}
\caption{Examples of failure cases.}
    \vspace{-0.5em}
    \label{fig:fail}
\end{figure}

Although the proposed feature interpolation is able to learn meaningful spatiotemporal features in most cases, there are a few failure cases as presented in \cref{fig:fail}. For instance, our method has difficulty in recovering 3D structures when small objects in a sequence rapidly move (\cref{fig:fail} left) or when there exist dynamic regions that are not observed in the training sequence (\cref{fig:fail} right).


\section{Conclusion}

In this paper, we propose a simple yet effective feature interpolation method for training dynamic NeRFs. Both the neural representation and the grid representation showed impressive performance, and the smoothness term applied to the intermediate feature vectors further improves the performance. Since these methods are unrelated to the existing methods of modeling deformations or estimating scene flows, we believe that the proposed method suggests a new direction of training dynamic NeRFs. 

While the neural representation model shows high-quality rendering results owing to the representation power of neural networks, it requires hours of training and seconds of rendering which impose barriers to real-time applications. On the other hand, the grid representation is able to render dynamic scenes in less than a second after a few minutes of training, which makes it more practical for real-world applications. Both representations are mutually complementary, and investigating hybrid representations that take advantages of both representations would be an interesting research direction.

{\small
\bibliographystyle{ieee_fullname}

}

\clearpage

\appendix
\section{Network Details}
The detailed structures of the networks for the neural and grid representation are explained in this section.
\subsection{Neural Representation}
\begin{figure}[t]
  \centering
   \includegraphics[width=0.98\linewidth]{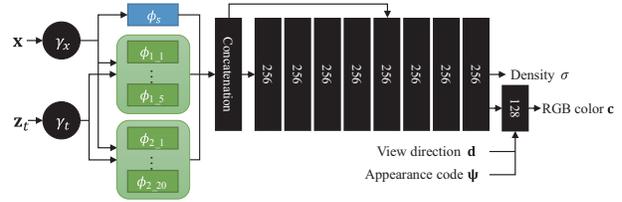}
   \caption{The network architecture of the neural representation. Input 3D position ($\mathbf{x}$) and the embedding vector for time $t$ ($\mathbf{z}_t$) are fed to the network after the positional encoding~\cite{mildenhall2021nerf} ($\gamma_x$ and $\gamma_t$). The template NeRF consists of 8 layers of MLP with hidden size of 256 and one additional layer for RGB color estimation.}
   \label{fig_supp:nn}
\end{figure}

The detailed network structure of the neural representation is illustrated in~\cref{fig_supp:nn}. Input 3D position $\mathbf{x}$ and the embedding vector $\mathbf{z}_t$ are encoded via the positional encoding used in~\cite{mildenhall2021nerf}. We set the maximum frequency levels of the positional encoding to 8 for $\mathbf{x}$ and 3 for $\mathbf{z}_t$. We adopted windowed positional encoding from~\cite{park2021hypernerf}, which weights the frequency bands of the positional encoding using a window function.

After the feature vector $\mathbf{v}(\mathbf{x},t)$ is extracted, it is fed into the template NeRF which consists of 8-layer MLPs with hidden size of 256 with ReLU activations, and one additional layer with hidden size of 128 for RGB color estimation. View direction and optional appearance code are also used as inputs for the RGB color estimation. We used the appearance code only for the DyNeRF dataset where an embedding vector is assigned to each camera.

Our implementation of the neural representation is based on the code from~\cite{park2021hypernerf} which is built using JAX~\cite{jax2018github}.

\subsection{Grid Representation}
\begin{figure}[t]
  \centering
   \includegraphics[width=0.95\linewidth]{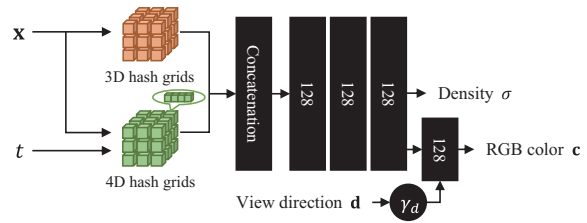}
   \caption{The network architecture of the grid representation. The extracted feature vector is fed to the template NeRF that has 3 layers of MLP with hidden size of 128 and one additional layer for RGB color estimation. View direction $\mathbf{d}$ is encoded using spherical harmonics as in~\cite{mueller2022instant}.}
   \label{fig_supp:grid}
\end{figure}

As depicted in~\cref{fig_supp:grid}, the NeRF MLP of the grid representation consists of 3-layer network with hidden size of 128 for density estimation, and one additional layer with hidden size of 128 for RGB color estimation. View direction is encoded using spherical harmonics of degree 4 following the implementation of \cite{mueller2022instant}. ReLU activations are used in all layers.

Our implementation of the grid representation is based on the code from~\cite{mueller2022instant} which is implemented using C++ and CUDA.

\section{Training Details}
Hyperparameter settings and training details for each dataset are described in this section. For all experiments, network parameters are optimized using ADAM optimizer~\cite{kingma2014adam}.


\subsection{Neural Representation}
We maintained two template NeRFs for optimization as in~\cite{mildenhall2021nerf}, one of which is trained from the sampled points that uses stratified sampling, and the other is trained using importance sampling for ray sampling strategy. We used 8 V100 GPUs or 4 A100 gpus to train the neural representation model. Each minibatch contains inputs that are sampled from 6,144 rays. It took approximately one day for training in D-NeRF and HyperNeRF datasets, and two days in DyNeRF datasets.

\noindent\textbf{D-NeRF} For D-NeRF datasets, initial learning rate is set to 0.002 and exponentially decayed to 0.0002 for 300,000 iterations. All training images are resized to $400\times400$ following the implementation of~\cite{pumarola2021d}.  The smoothness weight $\lambda$ is set to 0.01.

\noindent\textbf{HyperNeRF} For HyperNeRF datasets, initial and final learning rate are set to 0.001 and 0.0001 respectively. The smoothness weight $\lambda$ is set to 0.001. Training images are scaled by 0.25 and 0.5 for \textit{vrig} dataset and \textit{interp} dataset respectively.

\noindent\textbf{DyNeRF} The training images are downsized to 1K resolution ($1000\times750$). We set the initial and final learning rate to 0.001 and 0.00001 respectively. The smoothness weight $\lambda$ is set to 0.01, and the network is optimized for 600,000 iterations.

\subsection{Grid Representation}
The initial learning rate is set to 0.01 and is multiplied by 0.33 for every 10,000 iterations starting 20,000 iterations. As in~\cite{mueller2022instant}, we maintain the occupancy grid which is used to speed up training and rendering. To save the occupancy information of a whole sequence to a single occupancy grid, we assign random time frame value in addition to the input 3D points when querying the occupancy of the grid. Moreover, we adjust the decay weight of the values in the occupancy grid to 0.99 and set the threshold for culling to 0.0001.

\section{Derivation of the Smoothness Term}
We elaborate on the derivation of the smoothness
term used in the grid representation, which is discussed in
Section 3.3 of the main text. Applying the smoothness constraint to adjacent frames, as in the neural representation, the smoothness term becomes
\begin{equation}
    \mathit{L}_{s} = \lVert \mathbf{v}_d (\mathbf{x}, t) - \mathbf{v}_d (\mathbf{x}, t+1) \rVert _2 ^2 .
\label{eq4}
\end{equation}
Here, we assume that $\mathbf{x}$ is located on the 3D grid point without loss of generality, and we assume that the time frames $t$ and $t + 1$ lie between two grid points of time $t_a$ and $t_b$ ($t_a < t_b$).
Then, the output feature vector becomes a linear interpolation of the feature vectors of the two grid points, \ie,
\begin{equation}
    \mathbf{v}_d (\mathbf{x}, t) = s h_4 (\mathbf{x}, t_a ) + (1-s) h_4 (\mathbf{x}, t_b ) ,
\label{eq5}
\end{equation}
where $s = \frac{t_b - t}{t_b - t_a}$. Changing $t$ from $t$ to $t+1$ makes the weight $s$ decrease. Let $\epsilon$ denote the decrease in the weight, i.e., 

\begin{equation}
    \frac{t_b-(t+1)}{t_b-t_a} = \frac{t_b - t}{t_b-t_a} - \frac{1}{t_b-t_a} = s - \epsilon.
\end{equation}
, where $\epsilon = \frac{1}{t_b-t_a}$.

Then, the feature vector at time $t+1$ can be calculated as
\begin{equation}
    \mathbf{v}_d (\mathbf{x}, t+1) = (s-\epsilon) h_4 (\mathbf{x}, t_a ) + (1-s+\epsilon) h_4 (\mathbf{x}, t_b ) .
\label{eq6}
\end{equation}
Substituting \cref{eq5} and \cref{eq6} to \cref{eq4} yields
\begin{equation}
    \mathit{L}_{s} = \epsilon^2 \lVert h_4 (\mathbf{x}, t_a ) - h_4 (\mathbf{x}, t_b ) \rVert _2 ^2 .
\label{eq7}
\end{equation}
When the grid resolution is fixed, $\epsilon$ is proportional to $\frac{1}{n_f}$. Hence, we can obtain the smoothness term as the following form:
\begin{equation}
    \mathit{L}_{s} = \frac{1}{{n_f}^2} \lVert h_4 (\mathbf{x}, t_a ) - h_4 (\mathbf{x}, t_b ) \rVert _2 ^2 .
\label{eq8}
\end{equation}
In the case that $t$ and $t+1$ do not belong to the same grid, similar derivation reach to the same conclusion as~\cref{eq8} except that $t_a$ and $t_b$ are not adjacent but the closest grid points that satisfies $t_a \leq t \leq t+1 \leq t_b$. In practice, imposing the smoothness term only on adjacent grids is enough to improve performance. We applied~\cref{eq8} to every grid point that is used for feature vector calculation during the training, so the smoothness term can be assigned to a single grid point multiple times in a single iteration.

\section{Evaluation Details}

LPIPS metric may vary depending on which backbone network is used. We reported LPIPS using VGGNet for D-NeRF dataset and AlexNet for DyNeRF dataset. FLIP is calculated using weighted median for the DyNeRF dataset. We inferred those settings by implementing previous works or using publicly available code, and comparing them with our method.

When evaluating \textit{interp} dataset, the feature vector $\mathbf{v}_t$ is interpolated instead of interpolating the embedding vector $\mathbf{z}_t$. Since temporal interpolation of feature vectors is repeatedly occurred during the training, it is natural to interpolate the feature vectors rather than the embedding vectors to generate features of intermediate frames. This strategy improves overall performance by 0.5dB in PSNR compared to the embedding vector interpolation strategy.

In all experiments, network models are optimized and evaluated per sequence except the \textit{flame\_salmon} sequence in DyNeRF dataset which is separated to four sequences of 100 frames to have the same frames with the other sequences in the dataset. We provide detailed per-scene quantitative results for each dataset. From  \cref{tab1} to \cref{tab3}, the per-sequence performance of the neural representation models are presented. \cref{tab4} represents the per-sequence results of the grid representation on D-NeRF dataset.

\begin{table}
\centering
\begin{tabular}{l|cccc}
Sequence      & PSNR  & SSIM   & LPIPS  & AVG    \\ \hline
HellWarrior   & 25.40 & 0.953 & 0.0682 & 0.0349 \\
Mutant        & 34.70 & 0.983 & 0.0226 & 0.0100 \\
Hook          & 28.76 & 0.960 & 0.0496 & 0.0237 \\
BouncingBalls & 43.32 & 0.996 & 0.0203 & 0.0040 \\
Lego          & 25.33 & 0.943 & 0.0413 & 0.0307 \\
T-Rex         & 33.06 & 0.982 & 0.0212 & 0.0112 \\
StandUp       & 36.27 & 0.988 & 0.0159 & 0.0074 \\
JumpingJacks  & 35.03 & 0.985 & 0.0249 & 0.0098 \\ \hline
Mean          & 32.73 & 0.974 & 0.0330 & 0.0142
\end{tabular}
\caption{Per-sequence quantitative results of the neural representation model on D-NeRF dataset.}
\label{tab1}
\end{table}

\begin{table}[]
\centering
\resizebox{0.47\textwidth}{!}{
\begin{tabular}{lcc|lcc}
\multicolumn{3}{c}{\textit{vrig}} & \multicolumn{3}{|c}{\textit{interp}} \\
Sequence  & PSNR & SSIM & Sequence  & PSNR  & SSIM \\ \hline
Broom      & 20.48                 & 0.685       & Teapot      & 26.53                 & 0.933            \\
3D Printer  & 20.38                 & 0.678      & Chicken  & 27.99                 & 0.940                \\
Chicken     & 21.89                 & 0.869      & Fist     & 29.74                 & 0.933              \\
Peel Banana & 28.87                  & 0.965      & Fist     & 29.74                 & 0.933             \\
   &                   &        & Slice Banana & 28.39                  & 0.923                \\
      &                   &        & Lemon     & 31.31                 & 0.948               \\ \hline
Mean & 24.35	 & 0.866 & Mean & 28.67	 & 0.940
\end{tabular}
}
\caption{Per-sequence results of the neural representation model on HyperNeRF dataset.}
\label{tab2}
\end{table}

\begin{table}[]
\centering
\begin{tabular}{l|ccc}
Sequence           & PSNR  & LPIPS  & FLIP   \\ \hline
coffee\_martini    & 27.48 & 0.1143 & 0.1456 \\
cook\_spinach      & 33.12 & 0.0699 & 0.1262 \\
cut\_roasted\_beef & 33.63 & 0.0695 & 0.1221 \\
flame\_salmon\_1   & 27.66 & 0.1127 & 0.1468 \\
flame\_salmon\_2   & 26.91 & 0.1239 & 0.1464 \\
flame\_salmon\_3   & 27.05 & 0.1191 & 0.1558 \\
flame\_salmon\_4   & 26.72 & 0.1410 & 0.1493 \\
flame\_steak       & 33.11 & 0.0560 & 0.1398 \\
sear\_steak        & 33.24 & 0.0576 & 0.1396 \\ \hline
Mean               & 29.88 & 0.0960 & 0.1413
\end{tabular}
\caption{Per-sequence quantitative results of the neural representation model on DyNeRF dataset.}
\label{tab3}
\end{table}

\begin{table}
\centering
\begin{tabular}{l|cccc}
Sequence      & PSNR   & SSIM   & LPIPS  & AVG    \\ \hline
HellWarrior   & 24.33 & 0.936 & 0.1088 & 0.0466 \\
Mutant        & 32.04 & 0.977 & 0.0374 & 0.0152 \\
Hook          & 27.63 & 0.949 & 0.0859 & 0.0322 \\
BouncingBalls & 34.52 & 0.973 & 0.0633 & 0.0154 \\
Lego          & 25.16 & 0.935 & 0.0618 & 0.0364 \\
T-Rex         & 31.21 & 0.974 & 0.0445 & 0.0176 \\
StandUp       & 33.29 & 0.983 & 0.0315 & 0.0125 \\
JumpingJacks  & 30.51 & 0.968 & 0.0590 & 0.0211 \\\hline 
Mean          & 29.84 & 0.962 & 0.0615 & 0.0230  
\end{tabular}
\caption{Per-sequence quantitative results of the grid representation model on D-NeRF dataset.}
\label{tab4}
\end{table}

\section{Ablation Studies}
\subsection{Neural Representation}
To find the optimal structure of the network, we conducted experiments on various settings of the neural representation models. We changed number of MLPs for feature extraction, number of levels, and application of the smoothness term. All models are tested on $3dprinter$ sequence of \textit{vrig} dataset, and the results are shown in~\cref{tab5}. Two-level architecture with $n_0 = 5$, $n_1 = 20$ showed the best performance. The performance becomes worse in three-level architectures. Thus, using large number of networks does not guarantee performance improvements. In the case that the time slot between MLPs is too small, the network is optimized in only a few frames which prevents from learning meaningful features while imposing additional computational burden.

\begin{table}[t]
\resizebox{0.47\textwidth}{!}{
\begin{tabular}{l|ccc|c|cc}
Method      & $n_0$ & $n_1$ & $n_2$ & Smooth & PSNR           & MS-SSIM        \\ \hline
NeRF + Time & -   & -   & -   & -      & 21.05          & 0.847          \\ \hline
Ours-NN     & 2   & -   & -   & X      & 20.64          & 0.839          \\
            & 2   & 5   & -   & X      & 21.06          & 0.849          \\
            & 5   & 20  & -   & X      & 21.15          & 0.850          \\
            & 5   & 20  & -   & O      & \textbf{21.73} & \textbf{0.864} \\
            & 2   & 10  & 25  & O      & 21.15          & 0.849          \\
            & 5   & 20  & 50  & O      & 21.12          & 0.848         
\end{tabular}
}
\caption{Ablation studies on network architectures of the neural representation. \textit{3dprinter} sequence from $vrig$ dataset is used for evaluation.}
\label{tab5}
\end{table}

\subsection{Grid Representation}
\begin{figure}[t]
  \centering
   \includegraphics[width=0.9\linewidth]{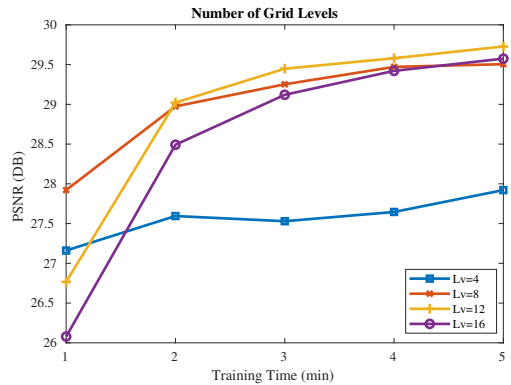}
\vspace{-0.5em}
   \caption{The performance of the grid representation model when different number of levels are used.}
   \label{fig_supp:grid1}
\end{figure}

\begin{figure}[t]
  \centering
   \includegraphics[width=0.9\linewidth]{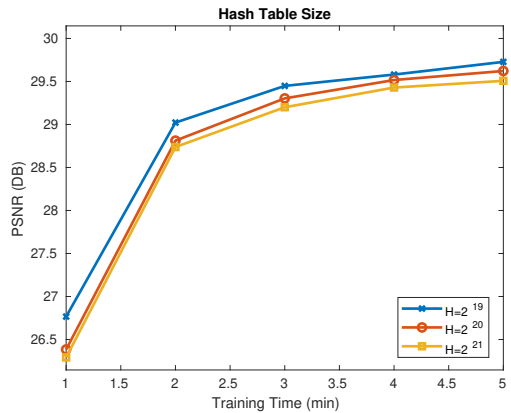}
\vspace{-0.5em}
   \caption{The performance of the grid representation model with various hash table size.}
   \label{fig_supp:grid2}
\end{figure}

We examined the effect of hash table size and number of grid levels. When small number of grid levels are used, as showin in~\cref{fig_supp:grid1}, the model converges faster, but shows slightly worse performance. Optimal performance is achieved when number of levels are set to 12. On the other hand, the performance tends to degrade when large hash table is used as observed in~\cref{fig_supp:grid2}. Not only for slow training speed, large hash table seems also inefficient to learn compact representations of the scene although further study including various real-scene data would be needed.



\begin{figure*}[t]
  \centering
  \hfill
  \begin{subfigure}{0.24\linewidth}
     \includegraphics[width=0.95\linewidth]{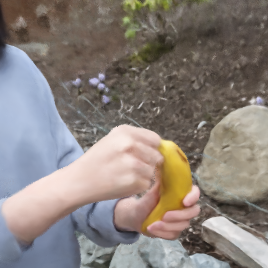}
    \caption{dynamic only}
    \label{fig:dynamic_only_banana}
  \end{subfigure}
  \begin{subfigure}{0.24\linewidth}
     \includegraphics[width=0.95\linewidth]{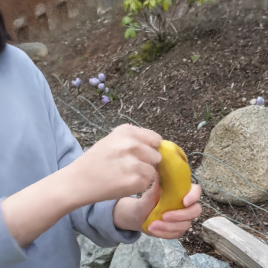}
    \caption{dynamic+static}
    \label{fig:ours_banana}
  \end{subfigure}
  \begin{subfigure}{0.24\linewidth}
     \includegraphics[width=0.95\linewidth]{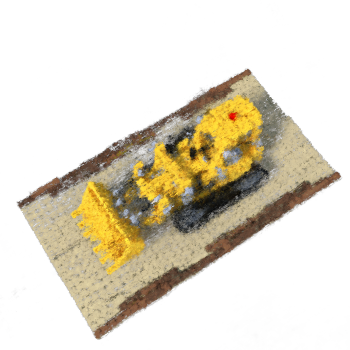}
    \caption{dynamic only}
    \label{fig:dynamic_only_lego}
  \end{subfigure}
  \begin{subfigure}{0.24\linewidth}
     \includegraphics[width=0.95\linewidth]{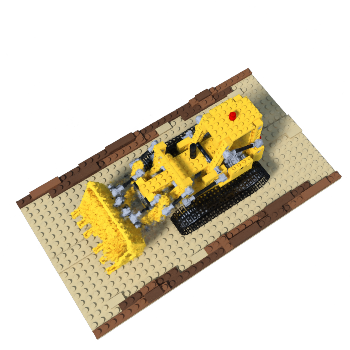}
    \caption{dynamic+static}
    \label{fig:ours_lego}
  \end{subfigure}
  \hfill
  \vspace{-0.3em}
  \caption{The effect of the static features on the neural representation ((a),(b)) and the grid representation ((c),(d)).}
  \label{fig:static}
\end{figure*}

\begin{figure*}[t]
  \centering
  \includegraphics[width=0.9\linewidth]{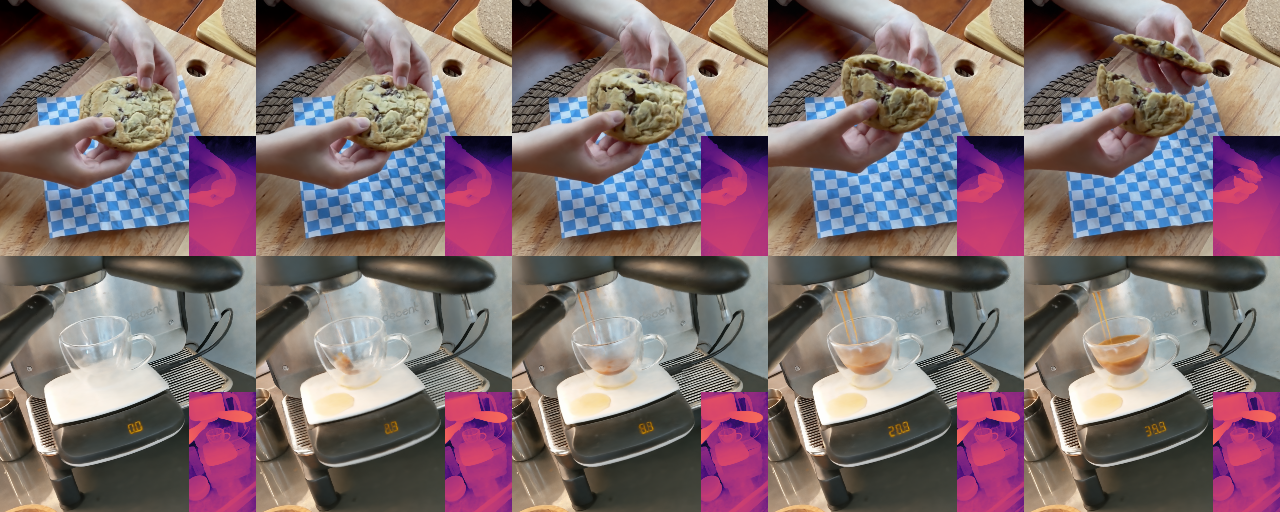}
  \vspace{-0.5em}
   \caption{Qualitative results of the neural representation model on the sequences containing topological variations.}
   \label{fig:topology}
\end{figure*}

\subsection{Effectiveness of the Static Features}
We showed qualitative results that validate the effectiveness of the static features in \cref{fig:static}. While one may think the quantitative improvement for the neural representation seems marginal, as it can be seen in \cref{fig:static} left, the model trained with static feature recovers fine details (\eg the texture of a stone and dirt). For the grid representation (\cref{fig:static} right), severe artifacts exist in static regions when only dynamic features are used.

\section{Additional Qualitative Results}
We conducted an additional experiment on the sequence containing significant topological variations, \textit{split-cookie} and \textit{espresso} from the HyperNeRF dataset. \Cref{fig:topology} shows qualitative results of the neural representation model on the sequence. We also included the depth estimation results at the corner of each image to ensure that 3D geometry is correctly estimated. Our method does not suffer from topological variations since it does not have any assumption about the shape topology, which verifies the flexibility of our approach.



\end{document}